\documentclass[lettersize,journal]{IEEEtran}
\usepackage{amsmath,amsfonts}
\usepackage{algorithmic}
\usepackage{algorithm}
\usepackage{array}
\usepackage[caption=false,font=normalsize,labelfont=sf,textfont=sf]{subfig}
\usepackage{textcomp}
\usepackage{stfloats}
\usepackage{url}
\usepackage{verbatim}
\usepackage{graphicx}
\usepackage{cite}
\begin{document}

\title{SpikeMM: Flexi-Magnification of High-Speed Micro-Motions}
\author{Baoyue Zhang$^{1,2\dag}$,Yajing Zheng$^{1*}$,Shiyan Chen$^{1}$,Jiyuan Zhang$^{1}$,Kang Chen$^{3}$,Zhaofei Yu$^{1*}$,Tiejun Huang$^{1}$\\
$^1$ School of Computer Science, Peking University\\
$^2$ Dalian University of Technology\\
$^3$ WuHan University\\
\texttt{yanyao\_715@163.com},\\
\texttt{\{strerichia002p,jyzhang\}@stu.pku.edu.cn,mrchenkang@whu.edu.cn} \\
\texttt{\{yj.zheng,yuzf12,tjhuang\}@pku.edu.cn}
\thanks{\textsuperscript{$*$} Corresponding authors.}
\thanks{\textsuperscript{$\dag$} Work done during an internship at Peking University.}
}




\maketitle

\begin{abstract}
The amplification of high-speed micro-motions holds significant promise, with applications spanning fault detection in fast-paced industrial environments to refining precision in medical procedures. However, conventional motion magnification algorithms often encounter challenges in high-speed scenarios due to low sampling rates or motion blur. In recent years, spike cameras have emerged as a superior alternative for visual tasks in such environments, owing to their unique capability to capture temporal and spatial frequency domains with exceptional fidelity. Unlike conventional cameras, which operate at fixed, low frequencies, spike cameras emulate the functionality of the retina, asynchronously capturing photon changes at each pixel position using spike streams. This innovative approach comprehensively records temporal and spatial visual information, rendering it particularly suitable for magnifying high-speed micro-motions.
This paper introduces SpikeMM, a pioneering spike-based algorithm tailored specifically for high-speed motion magnification. SpikeMM integrates multi-level information extraction, spatial upsampling, and motion magnification modules, offering a self-supervised approach adaptable to a wide range of scenarios. Notably, SpikeMM facilitates seamless integration with high-performance super-resolution and motion magnification algorithms. We substantiate the efficacy of SpikeMM through rigorous validation using scenes captured by spike cameras, showcasing its capacity to magnify motions in real-world high-frequency settings.
\end{abstract}


\section{Introduction}
\IEEEPARstart{M}{otion} magnification of high-speed micro movements is a promising technology with wide-ranging applications. It enables precise measurement and analysis of subtle motions within video, revealing imperceptible deformations and movements invisible to the naked eye. This capability holds significant implications for enhancing efficiency and safety across various industries. For instance, in industrial production, this technology can be employed for real-time fault detection in high-speed operational environments, aiding in the timely identification of mechanical failures, reducing downtime, and boosting productivity. 
\begin{figure*}[t]
    \centering
    \includegraphics[width=1.0\linewidth]{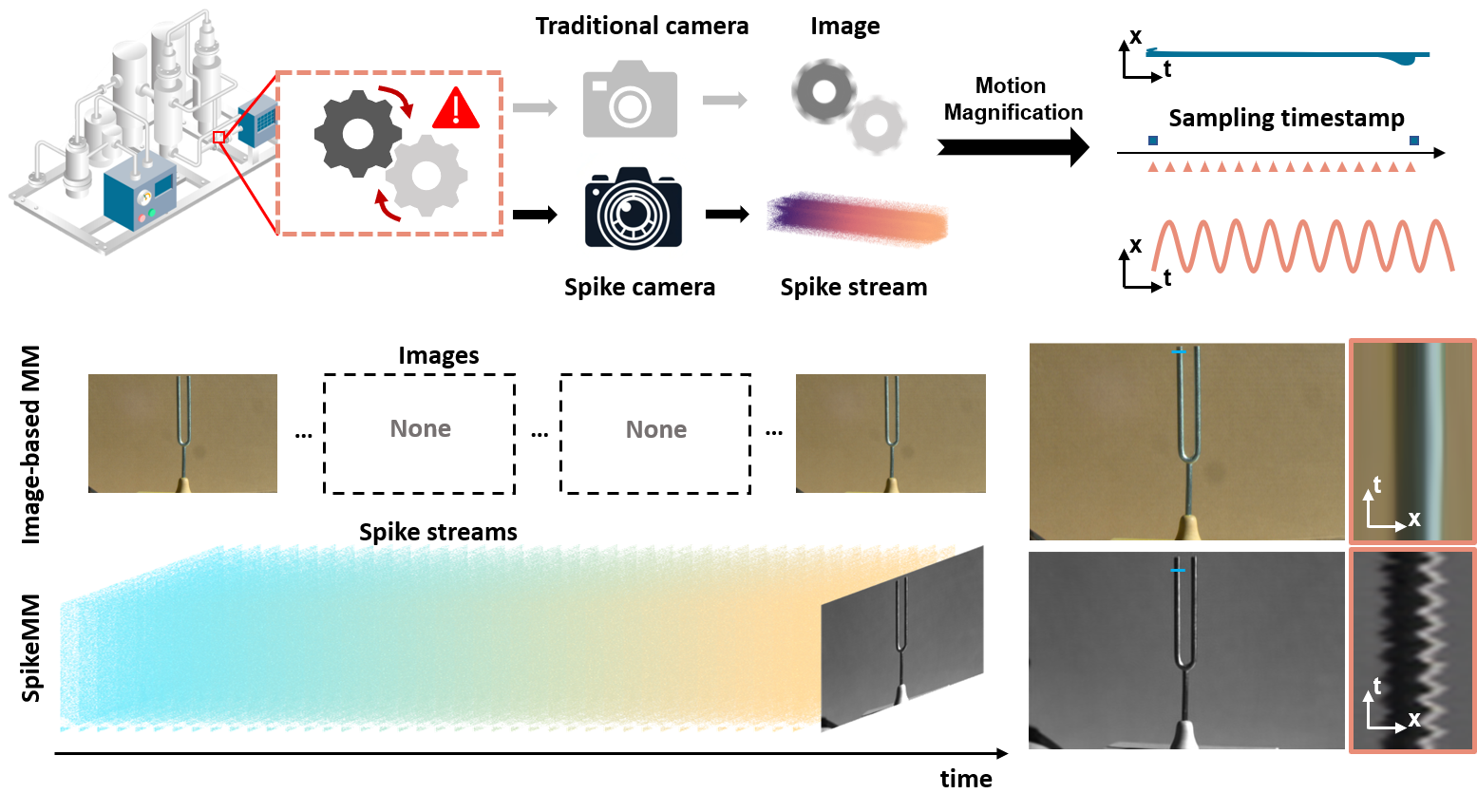}
    \caption{Comparison of traditional cameras and spike cameras in high-speed micro-motion amplification.}
    \label{fig:fig1}
\end{figure*}

Existing motion magnification techniques primarily rely on sequences of image frames \cite{oh2018learning,liu2005motion,singh2023lightweight,singh2023multi,pan2024self,zhang2017video,elgharib2015video,everingham2015pascal,flotho2022lagrangian,gao2022magformer,lin2014microsoft,wei2022novel,simoncelli1995steerable,wadhwa2013phase,wadhwa2014riesz,wang2022learning,wu2012eulerian}, amplifying target motions by analyzing subtle changes between consecutive frames. However, traditional cameras are limited by the exposure triangle, necessitating a trade-off between shutter speed and aperture size when capturing fast-moving objects. This often leads to blurry images, compromising the quality and detail of high-speed motions and thereby reducing the practicality and accuracy of motion magnification techniques. Additionally, the low output frequency of traditional cameras results in incomplete recording of high-frequency motions, rendering motion amplification inadequate in revealing minute deformations occurring at high speeds.
Amplifying motions in high-speed micro-movement scenes necessitates acquiring comprehensive spatiotemporal information. In recent years, spike cameras, which excel in high-speed imaging, have proven to be highly suitable \cite{STDP_zheng,tfp_tfi,bsn_chen,spk2img,zhu2020retina,red_chen,huang20221000,zhu2020retina,Laine2019,chen2023exploring,zhang2023learning,xiang2021learning,zhao2021super,hu2022optical,zhao2022learning,zhang2022spike,chen2023enhancing,zhang2023unveiling,xia2023svfi}. spike cameras~\cite{huang2022spiking}, inspired by the principles of the fovea centralis in the human eye's retina~\cite{yu2020toward,yan2020revealing,zheng2021unraveling}, utilize continuous spike flow to record changes in photons at each pixel position asynchronously. With their exceptionally high temporal sampling rates of 40,000 Hz, spike cameras can capture visual spatiotemporal information more comprehensively, enhancing the ability to capture and magnify high-speed micro-motions.

In this work, we propose the first self-supervised method utilizing spike cameras for motion magnification tasks, SpikeMM. There are three main challenges in employing spike cameras for amplifying high-speed micro-motions: (a) extracting spatiotemporal information from spike streams to simultaneously capture more motion details and scene textures, enhancing the effectiveness of motion magnification; (b) overcoming the compromise made in spatial resolution due to the pursuit of ultra-high temporal resolution in spike cameras, enabling the analysis of micro-motions at lower spatial resolutions; (c) as motion magnification inherently lacks ground truth (GT) in visual tasks, ensuring algorithm performance and effectiveness across various scenarios.

Our approach addresses these challenges through a self-supervised learning framework, composed mainly of multi-level information extraction, spatial upsampling, and motion amplification modules. In the multi-level information extraction module, we adopt a multi-level window length representation approach to input spike streams, overcoming the issues of motion blur caused by large windows and noise and discontinuity in motion resulting from small windows in previous methods \cite{tfp_tfi,bsn_chen}. Building upon the multi-level window feature representation, we further differentiate between moving and stationary pixels and input the feature fusion into the spatial upsampling module. In the spatial upsampling module, we employ Implicit Neural Representations (INR) to learn continuous function representations of data, capturing details with extreme precision and compression rates, achieving self-supervised super-resolution of spike information at any scale, thus enabling flexible magnification of high-frequency motions. Motion magnification primarily encompasses three methodologies: Eulerian Magnification, Lagrangian Motion Magnification, and Learning-based Magnification. SpikeMM serves as a plug-and-play module for Spike motion magnification. It can be flexibly integrated with these three motion magnification algorithms, effectively amplifying high-speed micro-motions. 

To underscore the effectiveness of spike cameras in magnifying high-speed micro-motions, the SpikeMM module will utilize a learning-based motion amplification algorithm to validate differences in performance compared to traditional videos in complex scenes.
We collected various high-speed motion scenes using spike cameras and tested the SpikeMM algorithm on these real-world scenarios. The results demonstrate that SpikeMM excels in magnifying micro-motions in high-speed scenes, effectively improving the accuracy and applicability of motion magnification technology. This breakthrough not only paves the way for the development of high-speed micro movement magnification techniques but also provides strong support for technological advancements in related industries.

The main contributions of this work are:
\begin{itemize}
\item[$\bullet$] Propose the first self-supervised spike-based framework for motion amplification of high-speed micro-movements. 
\item[$\bullet$] Introduce a multi-level information extraction module for spikes to balance motion blur and video consistency issues and leverage implicit neural representations of spike streams to enhance the model's ability to scale multi-level fusion features at any scale.
\item[$\bullet$] Construct the first spike stream dataset for motion magnification and validate the ability of SpikeMM to capture and amplify motion in high-frequency scenes.
\end{itemize}

\section{Related Works}
\subsection{Video Motion Magnification}
Motion magnification techniques, pivotal in enhancing the visibility of subtle movements within image sequences, are broadly classified into four distinct categories: phase-based\cite{wadhwa2013phase, wadhwa2014riesz}, Eulerian-based\cite{wu2012eulerian}, Lagrangian-based motion magnification\cite{liu2005motion}, and deep learning-based motion amplification algorithms\cite{oh2018learning, takeda2019video, singh2023multi, singh2023lightweight,pan2024self,wei2022novel}. Phase-based motion magnification methods\cite{wadhwa2013phase, wadhwa2014riesz} excel by leveraging the phase variation of each point in an image sequence to detect and accentuate motion. This approach, celebrated for its precision in phase information calculation, is exceptionally adept at amplifying minute, periodic motions such as those found in biological signals including heartbeat and respiration rates.
Eulerian-based motion magnification techniques\cite{wu2012eulerian}, on the other hand, concentrate on the temporal variations in pixel intensity. By amplifying these fluctuations, the technique significantly enhances the detection of small movements, making it particularly useful for observing non-periodic changes such as structural vibrations or variations in skin blood flow.
Lagrangian-based motion magnification strategies\cite{liu2005motion} distinguish themselves by focusing on the minute movements\cite{flotho2022lagrangian} of specific features or objects within the image. By analyzing and magnifying the dynamic changes of targeted objects, these methods prove invaluable for applications like tracking eye movements or the subtle displacements of mechanical components. Despite their effectiveness, these strategies rely heavily on robust feature detection algorithms to ensure continuity and accuracy in tracking, facing challenges in high-speed or blurry scenarios.

Recently, the advent of deep learning-based motion amplification algorithms\cite{oh2018learning, takeda2019video, singh2023multi, singh2023lightweight,pan2024self,wei2022novel} has introduced capabilities to process more complex scenes and motion types. 
However, despite their success, they encounter notable limitations in high-speed environments due to motion blur and the inherent sampling rate limitations of traditional cameras. Motion blur, resulting from rapid movements, hampers the analysis of subtle variations between frames, undermining the effectiveness of motion magnification by complicating feature point detection and tracking. Similarly, the fixed, often low, sampling rates of conventional cameras inadequately capture the nuances of high-speed movements, potentially leading to incomplete or inaccurate motion amplification.

\subsection{Spike-based Vision Algorithm}
The integral sampling mechanism of spike cameras enables them to record motion information at an extremely high frequency of 40,000 Hz, which also allows them to capture a wealth of textural information~\cite{huang2022spiking}. 
However, the spike stream is not visually friendly to humans, thus the reconstruction task stands as the most fundamental and crucial task for spike cameras.
Spike-based image reconstruction algorithms can be categorized into statistics-based methods~\cite{tfp_tfi, zhao2020high}, bio-inspired methods~\cite{STDP_zheng, zheng2023capture}, and deep learning-based methods~\cite{spk2img,zhang2023learning, bsn_chen}. Statistics-based spike high-speed imaging methods~\cite{tfp_tfi, zhao2020high} operate on the principle that pixel values are directly proportional to the rate of spike emission. These methods require a predefined window size for statistical spikes, making them sensitive to the trade-off between motion blur and noise.
Zhang \textit{et al.}\cite{zhang2023learning} proposed a wavelet-based representation to improve supervised reconstruction algorithms. Both Zhang \textit{et al.}\cite{zhang2023learning} methods necessitate synthetic datasets for network training. To mitigate data influence on training networks, Chen \textit{et al.}~\cite{bsn_chen,red_chen}, successfully recovered high-quality images from spike streams in a self-supervised manner.
To fully exploit the ultra-high-speed characteristics of spike cameras, researchers have demonstrated their unique advantages. For example, utilizing spike streams for tasks such as frame interpolation~\cite{xia2023svfi} and deblurring~\cite{chen2023enhancing} in RGB images, or employing dense spike streams for obstruction removal~\cite{zhang2023unveiling}.

The characteristics of spike cameras have also led to their application in high-speed, high dynamic range imaging of fast scenes~\cite{han2020neuromorphic,zhou2020unmodnet}. 
Researchers have explored spike cameras' potential in various tasks~\cite{zheng2023spikecv}, including image super-resolution~\cite{xiang2021learning,zhao2021super, zhao2023learning}, video frame interpolation~\cite{xia2023svfi}, optical flow estimation~\cite{hu2022optical,zhao2022learning,red_chen}, depth estimation~\cite{zhang2022spike}, high-speed object tracking and recognition~\cite{9985998, zhao2023spireco,li2022retinomorphic,zhu2022fpga}.
In this paper, we delve into the high-speed attributes of spike streams, pioneering their application in motion magnification tasks for the first time.

\section{Spike-based Motion Magnification}
\subsection{Preliminary}
\subsubsection{Spike Firing Mechanism}
\begin{figure}[ht]
    \centering
    \includegraphics[width=1.0\linewidth]{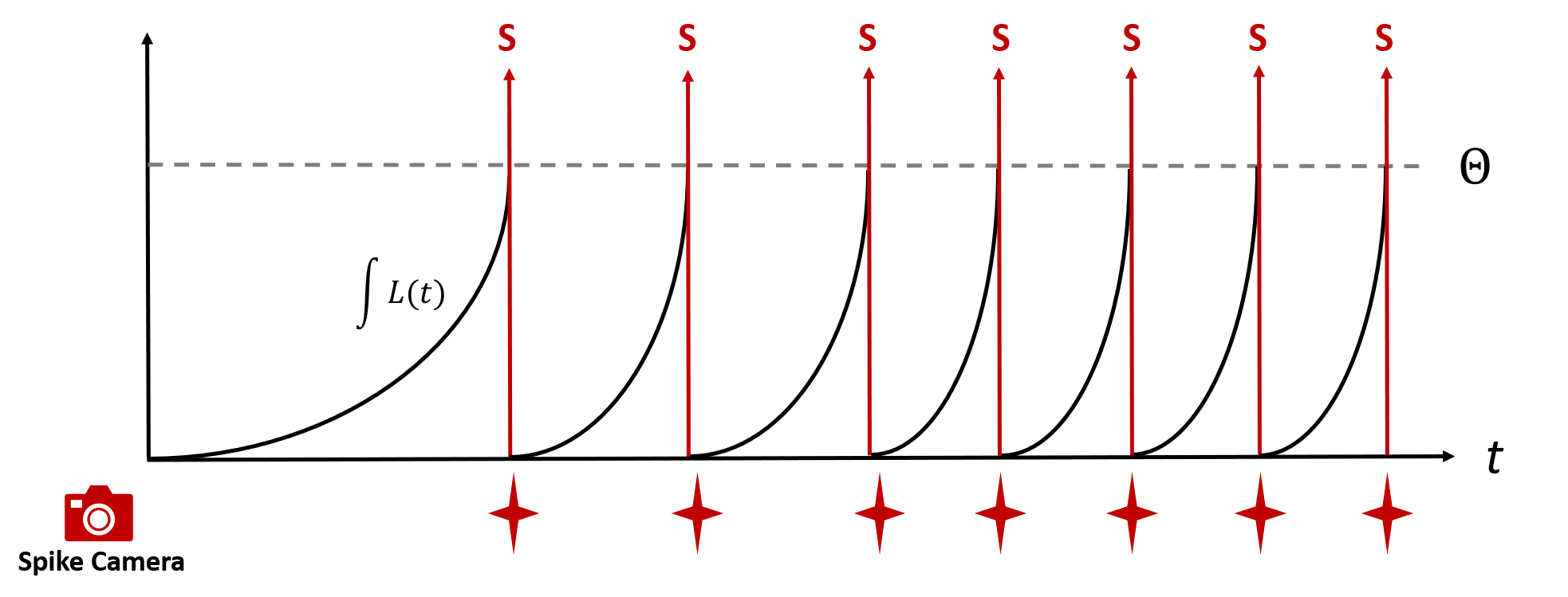}
    \caption{Illustration of the process by which the spike camera integrates light intensity and generates spikes, where red S denotes the moment of spike emission.}
    \label{fig:fig11}
\end{figure}

The spike camera mimics the sampling mechanism of the fovea of mammalian retinas, operating through an integrative sampling process~\cite{huang20221000}. Fig.~\ref{fig:fig11} shows the working principle of the spike camera. The spike camera emits a spike when the cumulative light intensity surpasses a certain threshold. 
Each pixel is equipped with an integrator that continuously records the incoming light intensity $L(t)$. When the accumulated light intensity in the integrator from the last spike point at time $t_p$ to a certain moment $t_0$ exceeds a preset threshold $\Theta$, the corresponding pixel will emit a spike signal and reset the accumulated light intensity in the integrator to zero. The mathematical expression is as follows:
\begin{equation}\int_{t_p}^{t_0}L(t)dt \geq \Theta\label{equ:spike_generation}.
\end{equation}

\subsubsection{Problem Statement}
\begin{figure*}[t]
    \centering
    \includegraphics[width=1.0\linewidth]{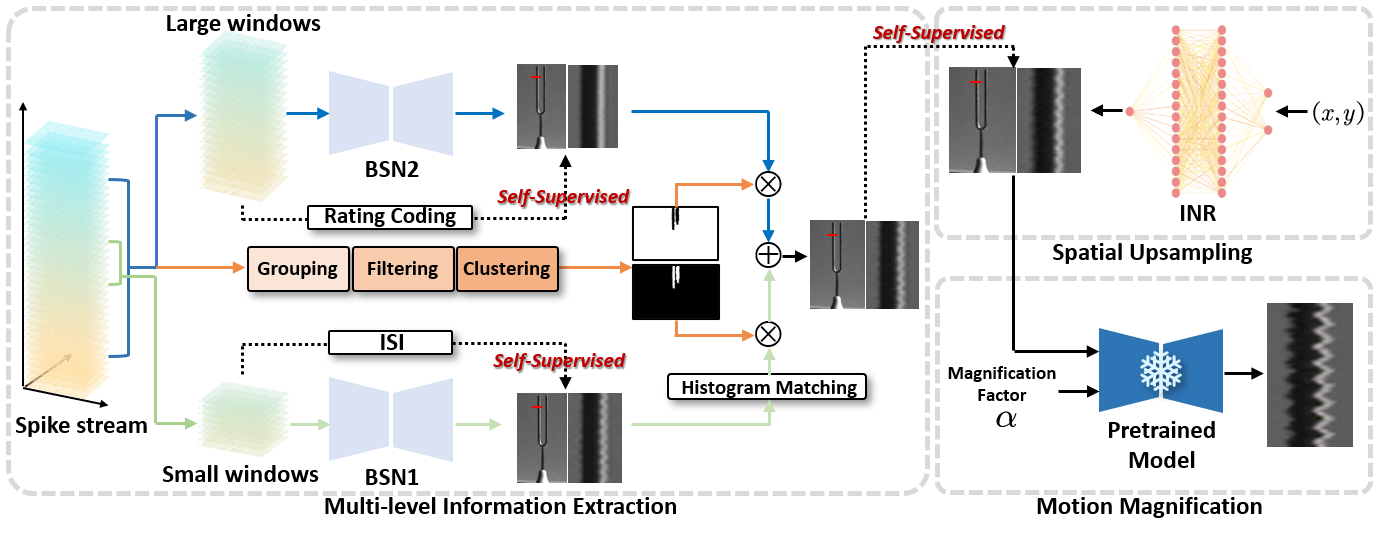}
    \caption{Architecture of the SpikeMM.}
    \label{fig:fig3}
\end{figure*}

Motion magnification is essentially a visual task without ground truth. In order to enable spike-based algorithms to perform well in various complex scenarios, our design of models primarily considers self-supervised learning to extract motion information from the spike flow. Subsequently, this information is provided to learning-based video motion magnification methods for inference and amplifying motion. 
The structure of the spike-based motion magnification algorithm we propose, \textit{SpikeMM}, is illustrated in Fig.~\ref{fig:fig3}. Furthermore, precise capture of motion information is required during video motion magnification to make the magnified motion more continuous and smooth. To achieve this effect, it is necessary to simultaneously consider obtaining more motion details while ensuring more scene textures. Therefore, in the initial processing of spike flow information in SpikeMM, we consider a multi-level information extraction approach for the spike flow. Long windows and short windows are utilized to extract information with different focuses, ultimately outputting a spike image sequence capturing complete motion information with video consistency. Building upon this, to compensate for the spatial resolution disadvantage of high-speed cameras and improve the utilization of detail information by the motion magnification algorithm, we adopt the implicit neural representation (INR) of spike data for arbitrary spatial upsampling. 

In the following, we will mainly describe how to perform multi-level information extraction on spike flow in Sec.~\ref{sec:Multi-level Information Extraction} and conduct spatial upsampling in Sec.~\ref{sec:Spatial Upsampling}.

\subsection{Multi-level Information Extraction}\label{sec:Multi-level Information Extraction}
\subsubsection{Representation of Spikes}
\begin{figure}[h]
    \centering
    \includegraphics[width=1\linewidth]{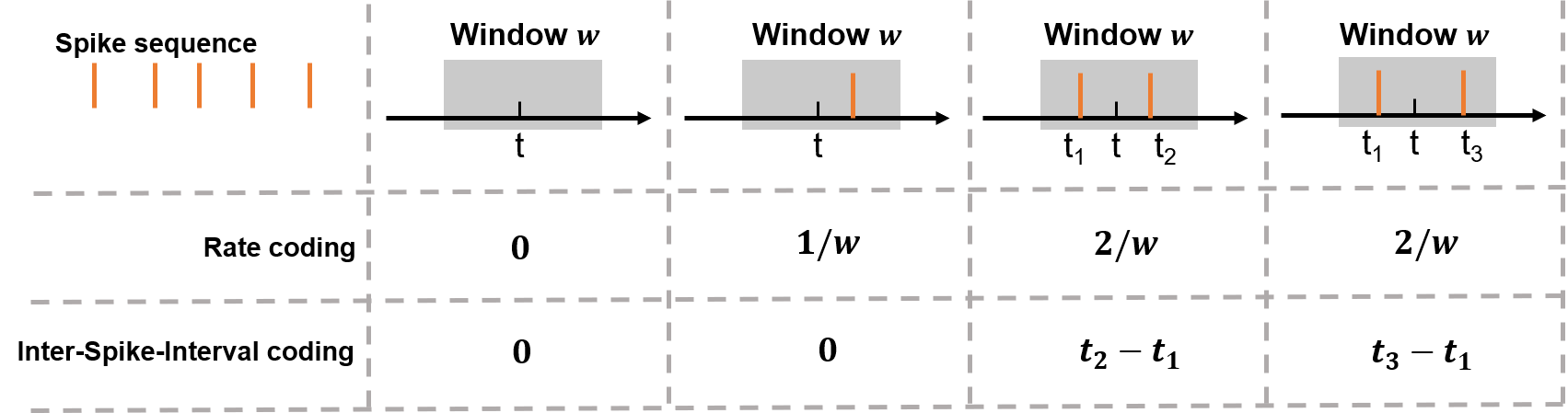}
    \caption{Examples of two encoding methods of spike sequence.}
    \label{fig:spike_coding}
\end{figure}

While the spike stream serves as a continuous signal representation, encoding information for network processing necessitates encoding at every moment. Presently, two prevalent spike encoding methods are in use. One method involves rate encoding \cite{tfp_tfi}, while the other utilizes inter-spike-interval (ISI) encoding \cite{tfp_tfi}. Both methods entail selecting a fixed window length to acquire the representation of the spike sequence S at each moment $t$. Fig.~\ref{fig:spike_coding} illustrates these two spike encoding methods.

Frequency encoding of spikes involves representing the frequency of spikes at a given moment t by the ratio of the number of spikes $N_w$ within a sliding time window. For the spike frequency at time $t$, its representation is given by:
\begin{equation}
\mathcal{R}_{t}=\frac{N_w}{w},
\end{equation}
where $w$ is the length of the window, which starts from $0$ and extends to $T$, with $t$ being the middle moment of this interval.

The spike interval encoding at time \( t \), involves locating the spikes before and after \( t \) within the window \( w \) centered at \( t \). If two spikes $t^+,t^-$ are found before and after \( t \), the ISI is calculated as the difference between the times of these two spikes. If fewer than two spikes are found, the ISI is considered as 0. This process can be formalized as follows:
\begin{align}
    ISI_t = \left \{
    \begin{array}{cc}
        t^+ - t^-, & \text{if} ~ \exists \{t^+,t^-\} \in [0,T], \\
        0, & \text{otherwise}.
    \end{array}
    \right.
\end{align}

These encoding methods offer unique advantages in processing spike information. Rate code enhances information representation with larger windows but is noise-sensitive in shorter ones, as shown in Fig.~\ref{fig:spike_coding}. Conversely, ISI thrives with shorter windows, improving sensitivity to spike variations, especially in moving pixels where motion information is crucial~\cite{STDP_zheng}. In video motion magnification, capturing subtle movements requires smaller windows to avoid motion blur but poses challenges in video consistency and may cause visual discontinuities due to the short window's limited texture information. Thus, for dynamic motion, short-window ISI is preferred for its accuracy without blur, while for static scenes or minor movements, long-window spikes are favored for comprehensive texture representation.

Therefore, in the Multi-level Information Extraction (MIE) module of spikes, we input spike streams with different window lengths and adopt the same approach as Chen \textit{et al.}~\cite{bsn_chen} to introduce the concept of Blind Spot Networks (BSN)~\cite{Laine2019,bsn_chen,chen2023exploring} to further eliminate noise from the spike flow. In BSN, the receptive field of each pixel does not include the pixel itself, forcing the network to reconstruct the intensity of the current pixel from spikes of surrounding pixels, thus avoiding learning the identity mapping of noise. In MIE, we simultaneously train two BSN networks to process spikes with two different window lengths.

In the BSN with short-window spikes $\mathbf{S}_{short}$, we will train the network, $\mathrm{BSN}_1$, in a self-supervised manner using the spike interval $ISI$ representation corresponding to that window length as pseudo-labels. The loss function $\mathcal{L}_{BSN1}$ is:
\begin{equation}
    \mathcal{L}_{BSN1}=\frac{1}{K}\sum_{k-1}^{K}\left\|\mathrm{BSN}_1(\mathbf{S}_{short})-\frac{1}{ISI}\right\|_2.
\end{equation}
In the long-window BSN, we will train the network, $\mathrm{BSN}_2$, using the spike frequency encoding of the long-window spikes $\mathbf{S}_{long}$ as pseudo-labels. Its loss function $\mathcal{L}_{BSN2}$ is:
\begin{equation}
    \mathcal{L}_{BSN2}=\frac{1}{K}\sum_{k-1}^{K}\|\mathrm{BSN}_2(\mathbf{S}_{long})-\mathcal{R}\|_2.
\end{equation}
\subsubsection{Feature Fusion}
\begin{figure*}[h]
    \centering
    \includegraphics[width=1\linewidth]{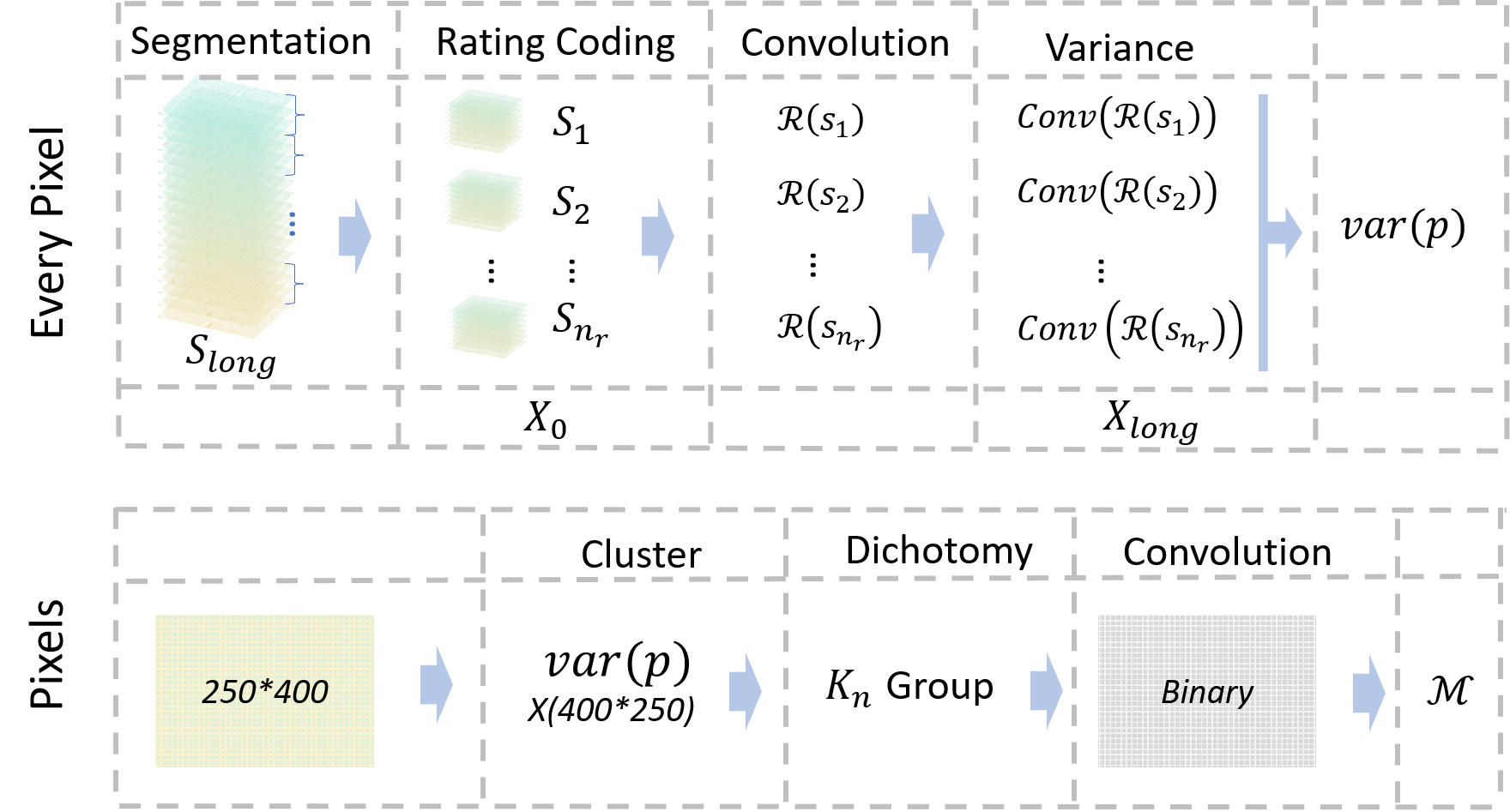}
    \caption{Pixels processing in feature fusion.}
    \label{fig:feature fusion}
\end{figure*}
The BSN1 and BSN2 models we developed are designed to capture motion cues from spike streams and maintain the continuity of texture details in videos, respectively. To leverage the unique characteristics of both models, we employed long-window spike streams as input to distinguish between regions with motion and stability. In the spike camera, each integrator within the sensor corresponds to a pixel in the spike stream data. The MIE is tasked with segmenting these pixels into ``stable points'', representing information from stable scenes in the sequence, and ``dynamic points'', representing dynamic information.
During the pixel segmentation process, we partition the long-window spike stream $S_{\text{long}}$ into \(n_r = w_l / w_s \) segments of spike streams, each having the same length as the short-window spike stream $S_{\text{short}}$. Here, \( w_l \) and \( w_s \) denote the window lengths of the long-window spike stream \( S_{\text{long}} \) and the short-window spike stream \( S_{\text{short}} \), respectively. This relationship can be expressed using the formula:
\begin{equation}
X_0=\left\{s_1, s_2, \ldots, s_{n_r}\right\}.
\end{equation}
Afterward, we separately perform frequency encoding on several segments of spike streams from \( X_0 \). These encoded segments are then passed through a convolutional layer to fuse spatial information. The mathematical expression for the obtained feature \( x_i \) is as follows:
\begin{equation}
    x_i=Conv(\mathcal{R}(s_i)).
\end{equation}
Thus, we obtain a new spatiotemporal representation \( X_{\text{long}} = \{ x_1, \cdots, x_{n_r} \} \) of the long-window spike stream. Due to the different fluctuation patterns in the frequency encoding of spike streams between dynamic and stable points, we compute the variance of \( Y \) at each pixel:
\begin{equation}
var(p) = \frac{1}{n_r - 1} \sum_{i=1}^{n_r} \left( X_{long}[p,i] - \overline{X_{long}[p]} \right)^2.
\label{eq:3}
\end{equation}
Subsequently, based on this variance, we perform K-means clustering and obtain \( K_n \) clusters. Up to this point, we have obtained sets of pixels with different stability levels, such as stable, dynamic, or moderately stable points. For unstable points, we consider them primarily generated by motion areas. We extract the motion regions based on the clustering results and then obtain a motion mask \( \mathcal{M} \) through a convolution operation. Using the mask derived from pixel classification, we merge information from both window sizes. BSN1 represents information in dynamic regions, while BSN2 represents information in static regions.

To better integrate information from both windows, before fusing the output values $\mathbf{O}_{BSN1}$ from BSN1, we adjust $\mathbf{O}_{BSN1}$ based on the results from BSN2, resulting in $\mathbf{O'}_{BSN1}$, where the transformation function \( \mathcal{T} \) is defined as:
\begin{equation}
 \mathcal{T} = \underset{\mathcal{T}}{\text{argmax}} \int_{-\infty}^{\infty} |h_{\mathbf{O'}_{BSN1}}(x) - h_{\mathbf{O}_{BSN2}}(y)| \, dy, 
\end{equation}
where \( h_{\mathbf{O'}_{BSN1}}(x) \) represents the histogram of the $\mathbf{O'}_{BSN1}$ after being transformed by the function \( \mathcal{T} \). $h_{\mathbf{O}_{BSN2}}(y)$ denotes the histogram of the output $\mathbf{O}_{BSN2}$ of BSN2.

The output fused feature \( \mathbf{O} \) of MIE is given by: 
\begin{equation}
   \mathbf{O}_{MIE} = \mathcal{M} \cdot \mathbf{O'}_{BSN1} + (1 - \mathcal{M}) \cdot \mathbf{O}_{BSN2}.
\end{equation}
\subsection{Spatial Upsampling}\label{sec:Spatial Upsampling}
Due to limited data bandwidth, trade-offs exist between temporal and spatial resolution in spike cameras. Existing spike camera sensors often retain low resolution (e.g., $250\times400$). To compensate for the lack of spatial resolution in spike cameras and enable them to flexibly perform motion magnification tasks in high-frequency, small-motion scenes, we propose introducing implicit neural representations (INRs) \cite{INR-de2023deep,INR-li2023regularize,INR-martel2021acorn,INR-ramasinghe2022beyond,INR-reiser2021kilonerf,INR-saragadam2022miner,INR-saragadam2023wire,INR-sitzmann2020implicit} to achieve arbitrary-scale upsampling. Specifically, we employ an MLP to perform the following mapping: $MLP: (x, y) \rightarrow c_{x,y}$. 
To better capture high-frequency details, we adopt the WIRE\cite{INR-saragadam2023wire}implementation.
We train the MLP in a self-supervised manner, with the training loss function given by:
\begin{equation}
\mathcal{L}_{sr} = || \mathbf{O}_{MIE}(x,y)-c_{x,y} ||_2^2.
\label{eq:inr_mlp}
\end{equation}
After training, by sampling superpixel grid positions $(x', y') \in (rH, rW)$, we can obtain super-resolution outputs $\mathbf{O}_{SR}$ at arbitrary scale $r$.
\subsection{Motion Magnification}
Our SpikeMM primarily verifies the effectiveness of spike representations over traditional images for the amplification of high-speed, subtle movements, and the flexible application of existing motion magnification algorithms. Furthermore, to leverage the efficacy of spike information representation in $\mathbf{O}_{MIE}$ and $\mathbf{O}_{SR}$ for the task of amplifying high-speed, subtle movements in complex scenes, we have employed the first learning-based motion magnification algorithm proposed by Oh \textit{et al.} \cite{oh2018learning}, along with the recent state-of-the-art learning-based method (Singh \textit{et al.} \cite{singh2023multi}.). Specifically, we input $\mathbf{O}_{SR}$ into the motion magnification algorithm, set a motion magnification factor $\alpha$, and obtain the final magnified image output $\Tilde{{\mathbf{I}}}$ from SpikeMM. 
This process can be expressed with the formula:\begin{equation}\Tilde{\mathbf{I}}(\mathbf{O}_{SR}, t) = f(\mathbf{O}_{SR} + (1+\alpha)\cdot\delta(\mathbf{O}_{SR},t)),
\end{equation}
where $\delta(\mathbf{O}_{SR}, t)$ denotes the displacement function at time $t$. The function \(f\) can be flexibly replaced with any currently mature motion magnification algorithm. 

\section{Experiments}
\subsection{Camera System}
\begin{figure}[h]
    \centering
    \includegraphics[width=0.6\linewidth]{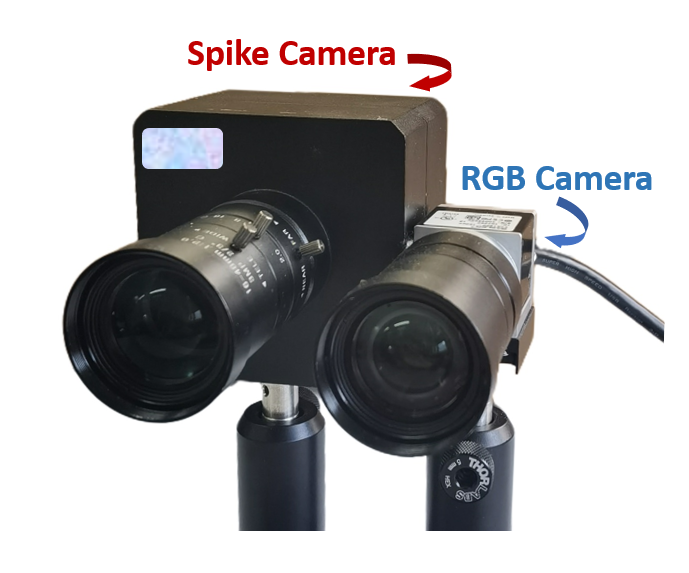}
    \caption{The camera system that we used to collect data.}
    \label{fig:fig12}
\end{figure}

As shown in Fig.~\ref{fig:fig12}, we employ a camera system comprising a spike camera and an RGB camera to simultaneously capture Spike and RGB data, facilitating a comparison of the Spike camera and RGB camera in identical scenes. Note that the rgb camera is used for comparison purposes and not as an input to SpikeMM, which only needs a spike camera.The details of the spike camera and the RGB camera are listed below.
\begin{itemize}
    \item \textbf{RGB camera} that we use is \textit{Basler acA1920-150uc}. Simultaneously capturing images with a Spike camera and an RGB camera allows us to intuitively compare the differences between them.
    \item \textbf{Spike camera} that we use is \textit{Spike Camera-001T-Gen2} with a spatial resolution ($H\times W$) of $250 \times 400$, and a temporal resolution of $20,000$ Hz. The output is a three-dimensional binary spike sequence $\mathrm{\mathbf{S}}=\{0,1\}^{H\times W \times T}$, where $T$ is the recording time duration.
\end{itemize}
\subsection{Dataset}
We constructed a hybrid system combining a spike camera and an RGB camera, through which we simultaneously captured spike and RGB data from four indoor scenes in the real world including `Tuning Fork', `Short Ruler', `Long Ruler', `Balloon'. Every scene has 100 periods of high-speed micro-motions. These scenes included objects with high-frequency minute motions, stationary objects, air current movements, and shadow changes caused by sunlight, as well as desktop vibrations induced by motion, \textit{etc}.

\subsection{Training Details}
\subsubsection{BSN}
As a fully self-supervised approach, we can train BSN with only spike sequences from a single scene. Specifically, we adopt the BSN implementation form in ~\cite{bsn_chen}, where the network overall follows a U-shaped structure and blind spots are constructed using shifted convolutions. The batch size is set to 1, and the spike stream is cropped to $256\times256$. The BSN is optimized using Adam optimizer with a learning rate of 2e-4 for 3000 iterations.

\subsubsection{INR}
We use WIRE~\cite{INR-saragadam2023wire} as the specific implementation of INR. The network has an input dimension of $2$ and an output dimension of $1$, with two hidden layers of dimension $256$. We set the batch size to the total number of pixels in the input image, with each batch representing a coordinate point in the grid. We use the Adam optimizer with a learning rate of 1e-3. And we train the WIRE for 3000 iterations per image. All experiments are completed on an RTX 2080 GPU.

\begin{figure*}[h]
    \centering
    \includegraphics[width=1.0\linewidth]{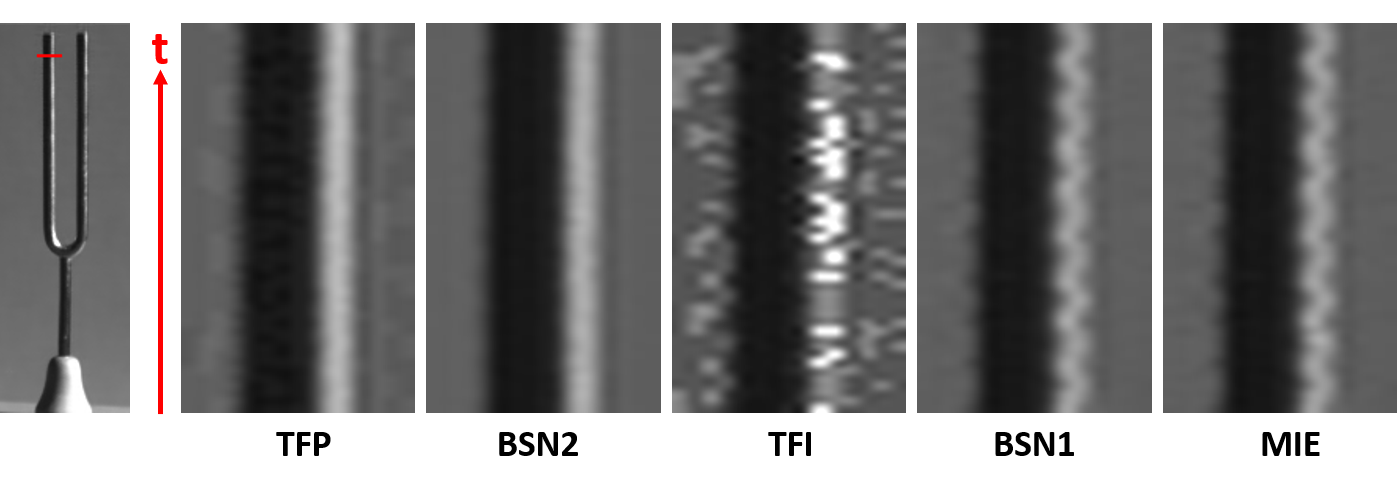}

    \caption{Temporal evolution of motion in `TuningFork' scene of different reconstruction methods in 21.6ms.}
    \label{fig:fig5}
\end{figure*}

\begin{figure*}[h]
    \centering
    \includegraphics[width=1.0\linewidth]{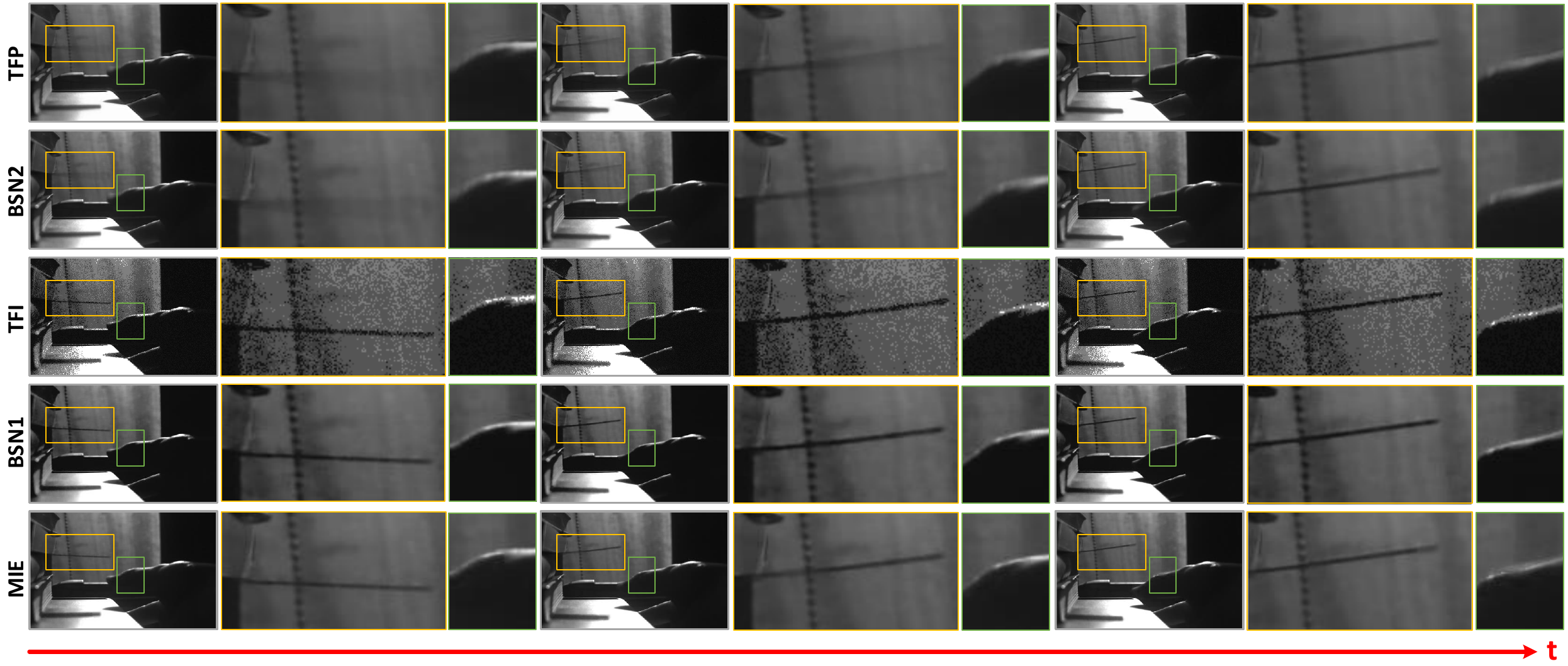}
  
    \caption{Image reconstruction results in the `LongRuler' sequence of different methods.}
    \label{fig:fig6}
\end{figure*}

\subsection{Evaluation of The Spike Representation}
The processing of motion blur in spike scenes is crucial for the motion magnification task. We employed various methods with differing window sizes to analyze the motion blur in our scenes. Qualitative results of `TuningFork' and `LongRuler' are shown in Fig.~\ref{fig:fig5} and Fig.~\ref{fig:fig6}, respectively. The results indicate that long-window methods result in more motion blur effects, which are unacceptable for motion magnification, while short-window methods perform better in terms of reducing motion blur.

\begin{figure}[h]
    \centering
    \includegraphics[width=1.0\linewidth]{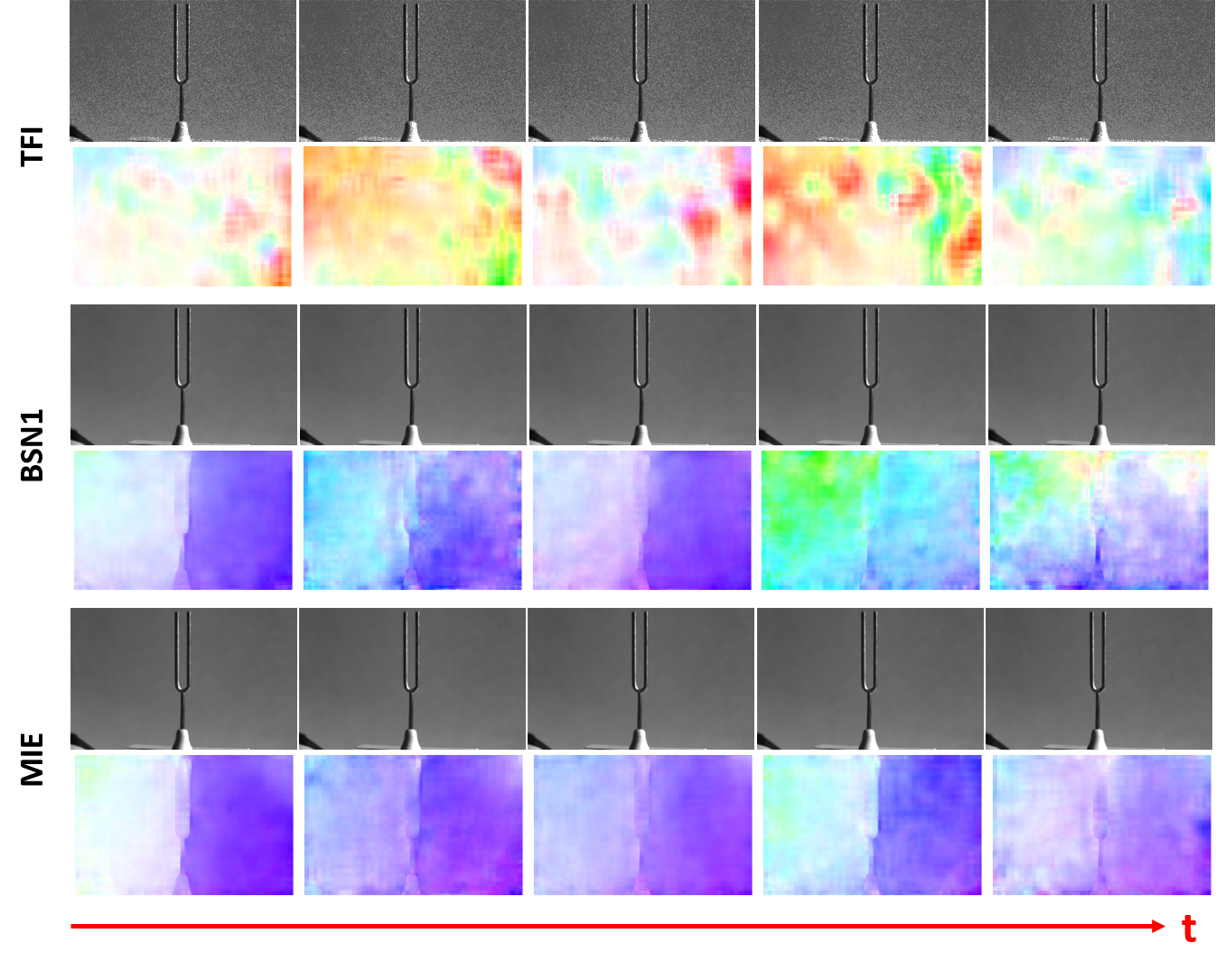}
    \caption{Examples of the optical flow results of the `TuningFork' based on different reconstruction methods.}
    \label{fig:fig7}
\end{figure}
\begin{figure}[h]
    \centering
    \includegraphics[width=1.0\linewidth]{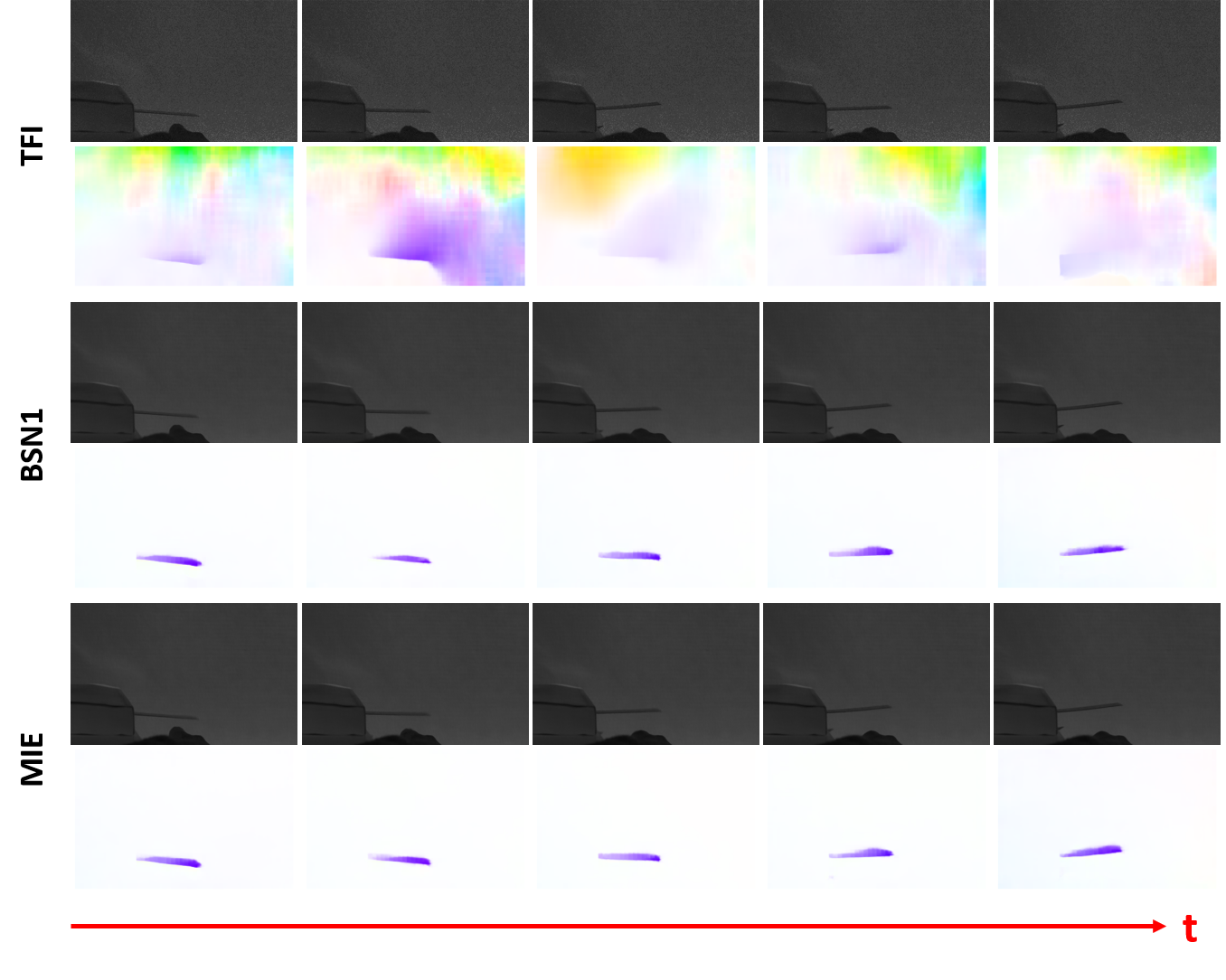}
    \caption{Examples of the optical flow results of `ShortRuler' based on different reconstruction methods.}
    \label{fig:fig8}
\end{figure}

For the video consistency of spike information results, we reflected the impact of noise by calculating the optical flow of the spike video stream with RAFT\cite{teed2020raft}. The optical flow results for the `Tuning Fork' and `Short Ruler' are presented in Fig.~\ref{fig:fig7} and Fig.~\ref{fig:fig8}. The optical flow findings indicate that MIE exhibits less noise influence in video manifestation.

We introduced two objective metrics, Flow Consistency, and Motion Smoothness, to evaluate the output spike video stream of the processed spike scenes.  According to our knowledge, this is the first video-level evaluation for spike scene information processing. Let $F_i$ be the optical flow vector between the $i^{th}$ frame and the $(i+1)^{th}$ frame. For the calculation of flow consistency, we first compute the difference $\Delta F_i=F_{i+1}-F_i$ in optical flow vectors between consecutive frames. Next, we calculate the standard deviation of the differences:
\begin{equation}
\mu=\frac{1}{N-1} \sum_{i=1}^{N-1}\left\|\Delta F_i\right\|, 
\end{equation}
\begin{equation}\sigma=\sqrt{\frac{1}{N-1} \sum_{i=1}^{N-1}\left(\left\|\Delta F_i\right\|-\mu\right)^2}, 
\end{equation}
where $N$ is the total number of frames, and $\left\|\Delta F_i\right\|$ is the magnitude of the difference in optical flow vectors for the $i^{th}$ frame. $\sigma$ indicates the consistency of changes in the optical flow vectors. A lower $\sigma$ suggests that the motion in the video is more visually coherent and smooth.
Motion smoothness mainly focuses on the variation of optical flow vectors over time. First, we calculate the magnitude of the optical flow vector for each frame $S_i=\left\|F_i\right\|$ where $S_i$ is the magnitude of the optical flow vector for the $i^{th}$ frame. Then:
\begin{equation}
\mu_S=\frac{1}{N} \sum_{i=1}^N S_i, 
\end{equation}
\begin{equation}\sigma_S=\sqrt{\frac{1}{N} \sum_{i=1}^N\left(S_i-\mu_S\right)^2}, 
\end{equation}
where $\sigma_S$ reflects the consistency of motion intensity. A lower $\sigma_S$ indicates that the motion in the video is smoother and more consistent.

The results of Flow Consistency and Motion Smoothness are shown in Table.~\ref{tab:table1} and Table.~\ref{tab:table2}. The results indicate that, compared with TFI and BSN1, MIE exhibits the best performance in terms of optical flow consistency and motion smoothness.
\begin{table}[!t]
\caption{Comparison of  Flow Consistency(\(\downarrow\)) on real-life spike streams.\label{tab:table1}}
\centering
\begin{tabular}{@{}l@{\hspace{30pt}}c@{\hspace{20pt}}c@{\hspace{20pt}}c@{}}
\hline
Name  & TFI\cite{tfp_tfi} & BSN1 \cite{bsn_chen}  & Proposed \\ 
\hline
Tuning fork   & 0.0360 & 0.0148   & \textbf{0.0066}  \\
Short ruler  & 1.0464 & 0.0048    & \textbf{0.0042}  \\
Long ruler   & 0.0854 & 0.0978 & \textbf{0.0318}  \\
Balloon & 0.3438 & 0.0066  & \textbf{0.0033} \\
\hline
Average & 0.3779 & 0.0310 & \textbf{0.0115}  \\
\hline
\end{tabular}
\end{table}
\begin{table}[!t]
\caption{Comparison of Motion Smoothness(\(\downarrow\)) on real-life spike streams.\label{tab:table2}}
\centering
\begin{tabular}{@{}l@{\hspace{30pt}}c@{\hspace{20pt}}c@{\hspace{20pt}}c@{}}
\hline
Name  & TFI\cite{tfp_tfi} & BSN1 \cite{bsn_chen}  & Proposed \\ 
\hline
Tuning fork   & 0.0530 &  0.0174   & \textbf{0.0103}  \\
Short ruler  & 0.8124 &0.0110    & \textbf{0.0043}  \\
Long ruler   & 0.0886 & 0.1213  & \textbf{0.0384}  \\
Balloon & 0.3204 & 0.0099  & \textbf{0.0077} \\
\hline
 Average & 0.3186 & 0.0339 & \textbf{0.0152}  \\
\hline
\end{tabular}
\end{table}
\subsection{Ablation Study}
\begin{figure}[h]
    \centering
    \includegraphics[width=1.0\linewidth]{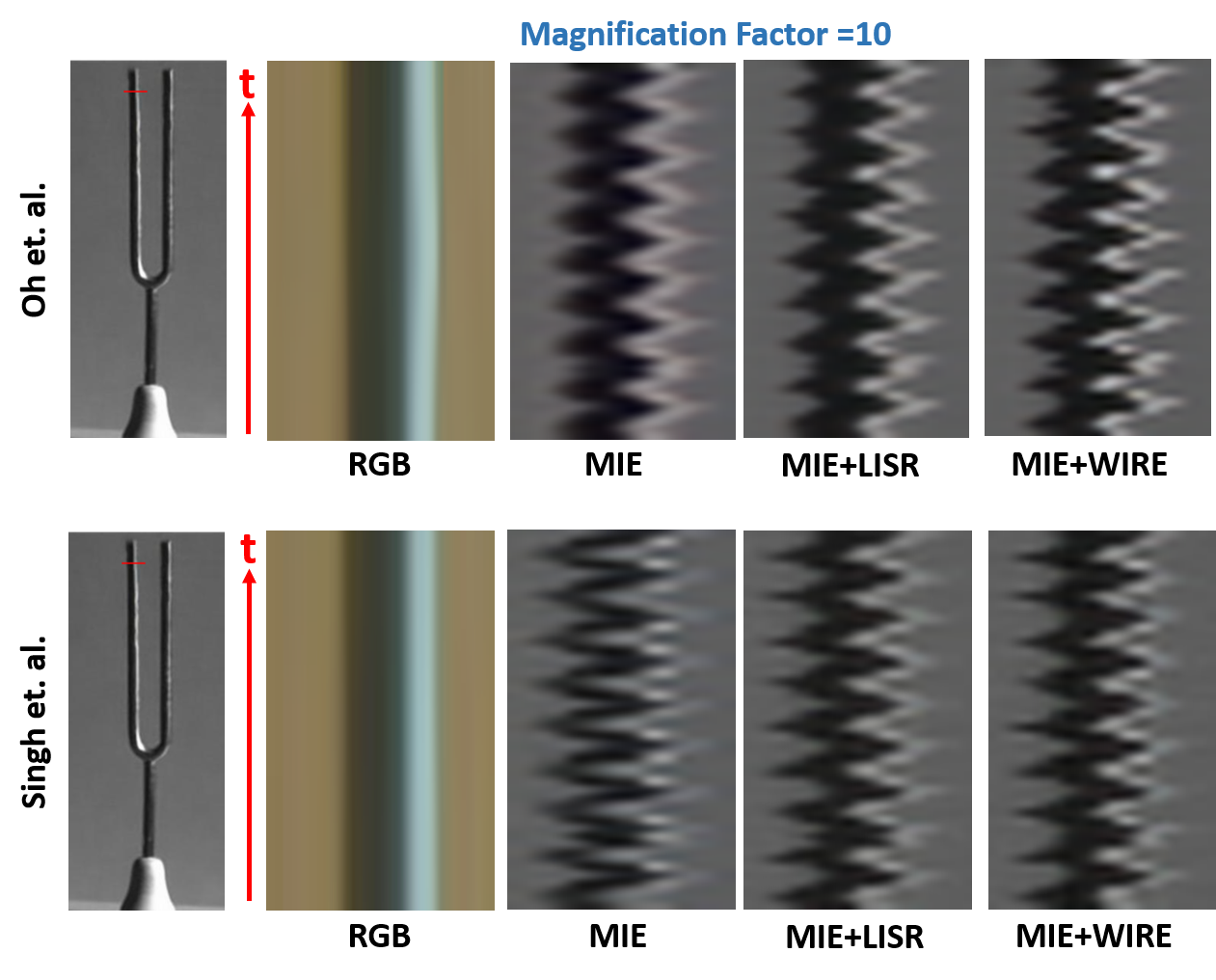}
    \caption{Comparison of qualitative results of `TuningFork' sequence ablation experiments in 21.6ms.}
    \label{fig:fig9}
\end{figure}

\begin{figure}[h]
    \centering
    \includegraphics[width=1.0\linewidth]{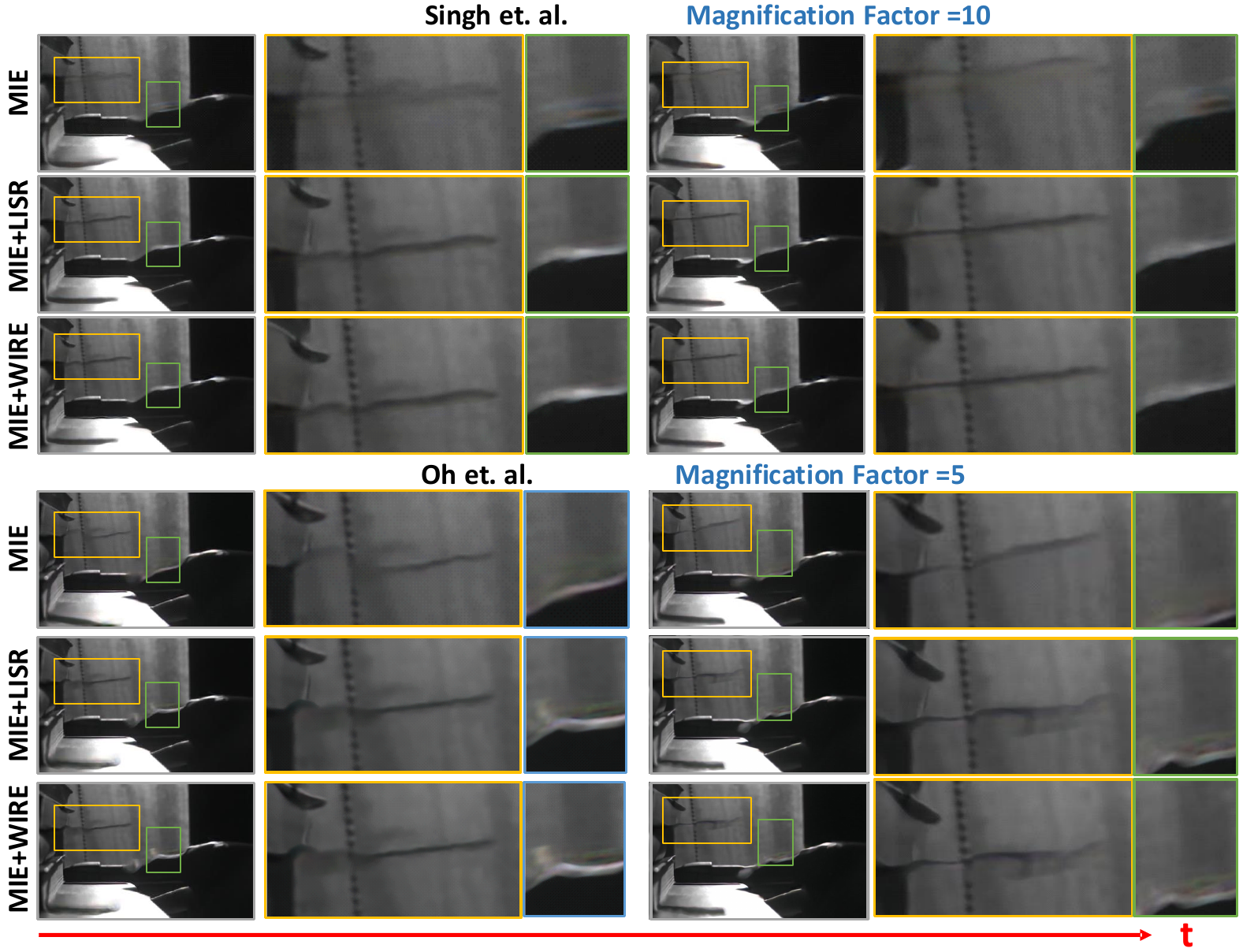}
    \caption{Ablation experiments on motion amplification effects under different super-resolution methods.}
    \label{fig:fig10}
\end{figure}

We conducted six sets of ablation experiments to demonstrate the effectiveness of each module, comparing the results of MIE, MIE post-Linear Interpolation super-resolution (LISR), and MIE post-INR (WIRE\cite{INR-saragadam2023wire}) in terms of magnification effects using both motion magnification method of Oh \textit{et al.}\cite{oh2018learning}. and motion magnification method of Singh \textit{et al.}\cite{singh2023multi}.
In ablation experiments with magnification factors of 10 in `Tuning Fork' at Fig.~\ref{fig:fig9} and factors of 5, 10 in `Long Ruler' scenes at Fig.~\ref{fig:fig10}, the results indicate that the combination of MIE and WIRE followed by the method of Oh \textit{et al.} and Singh \textit{et al.} shows the best performance. This is evident in (a) an increase in motion amplitude after super-resolution in the `Tuning Fork' scene compared to IME, which is without super-resolution, due to reduced motion blur. And image contrast is enhanced. Similarly, the reduction in motion blur is more pronounced in the `Long Ruler' scene; (b) compared to the LISR method, WIRE shows some improvement in motion amplitude and achieves higher image contrast; (c) both methods of motion magnification have demonstrated very good results. In comparison with the method of Singh \textit{et al.} method, the approach of Oh \textit{et al.} in the `Long Ruler' scene, achieves visually equivalent effects with a relatively smaller motion magnification factor.

Ablation experiments reveal that SpikeMM performs exceptionally well in motion magnification, demonstrating the flexibility to seamlessly integrate with various motion magnification methods. It has the potential to enable basic motion magnification techniques, like the method of Oh \textit{et al.}\cite{oh2018learning}, to rival, and possibly surpass the more complex methods. The IME module effectively converts spike stream data into frame sequence inputs, while the INR super-resolution module, by enhancing resolution, significantly aids in improving the magnification effects of motion magnification.
\subsection{Video Comparison}
We conduct motion magnification tasks in four different scenes and demonstrate the experimental results for each scene in the form of videos. In the following, some examples of qualitative comparison are given. For more specific details, please see our supplemental video. 
 In the scene of \textbf{`Tuning Fork',} we conduct comprehensive experiments on the effectiveness of SpikMM, including (1) comparative analysis of spike data processing methods between MIE and other approaches (TFI, TFP, BSN1, BSN2), and effects of applying them directly to motion magnification (MM) models \cite{oh2018learning,singh2023multi}; (2) comparison of the effects of MIE, $\mathcal{\mathbf{O}}_{MIE}$, and post-super-resolution MIE, $\mathcal{\mathbf{O}}_{SR}$, with different magnification methods.

\begin{figure}[h]
    \centering
    \includegraphics[width=1.0\linewidth]{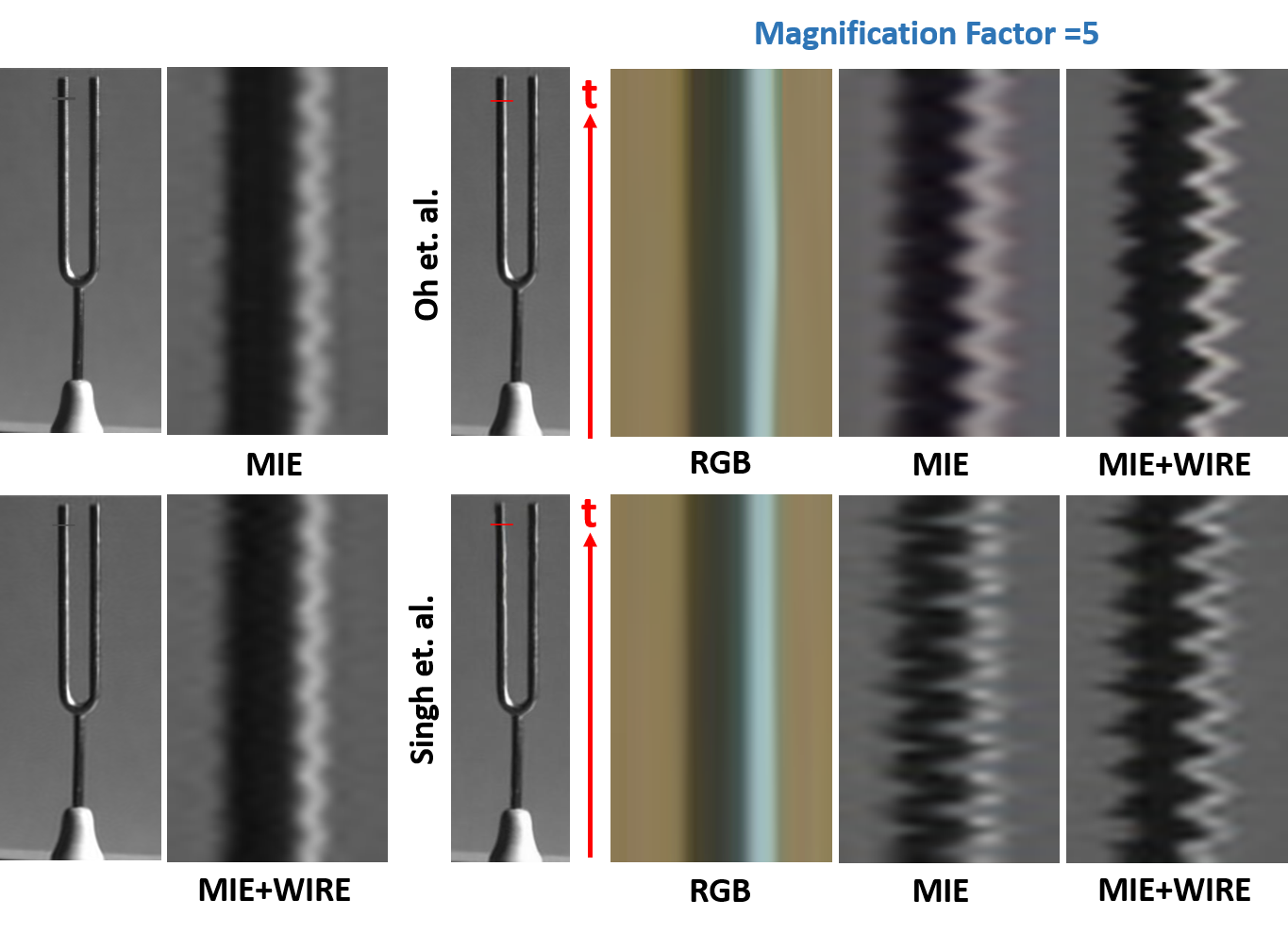}
    \caption{Temporal evolution of motion in `TuningFork' scene of different methods.}
    \label{fig:fig13}
\end{figure}
The temporal evolution of motion in the ‘Tuning Fork’ scene using different methods is displayed in Fig.~\ref{fig:fig13}.

For the scenes of \textbf{`Long Ruler'}, \textbf{`Short Ruler'} and \textbf{`Balloon'}, we conduct a comparison of different spike data processing methods and contrast the performance of MIE in motion magnification under various conditions. 
The video presentation of `Long Ruler' demonstrates that the MIE effectively maintains the continuity of the video while accurately capturing the motion states of high-frequency moving objects. Additionally, the spatial upsampling module can improve resolution and enhance the effect of motion magnification as well.

\begin{figure}[h]
    \centering
    \includegraphics[width=1.0\linewidth]{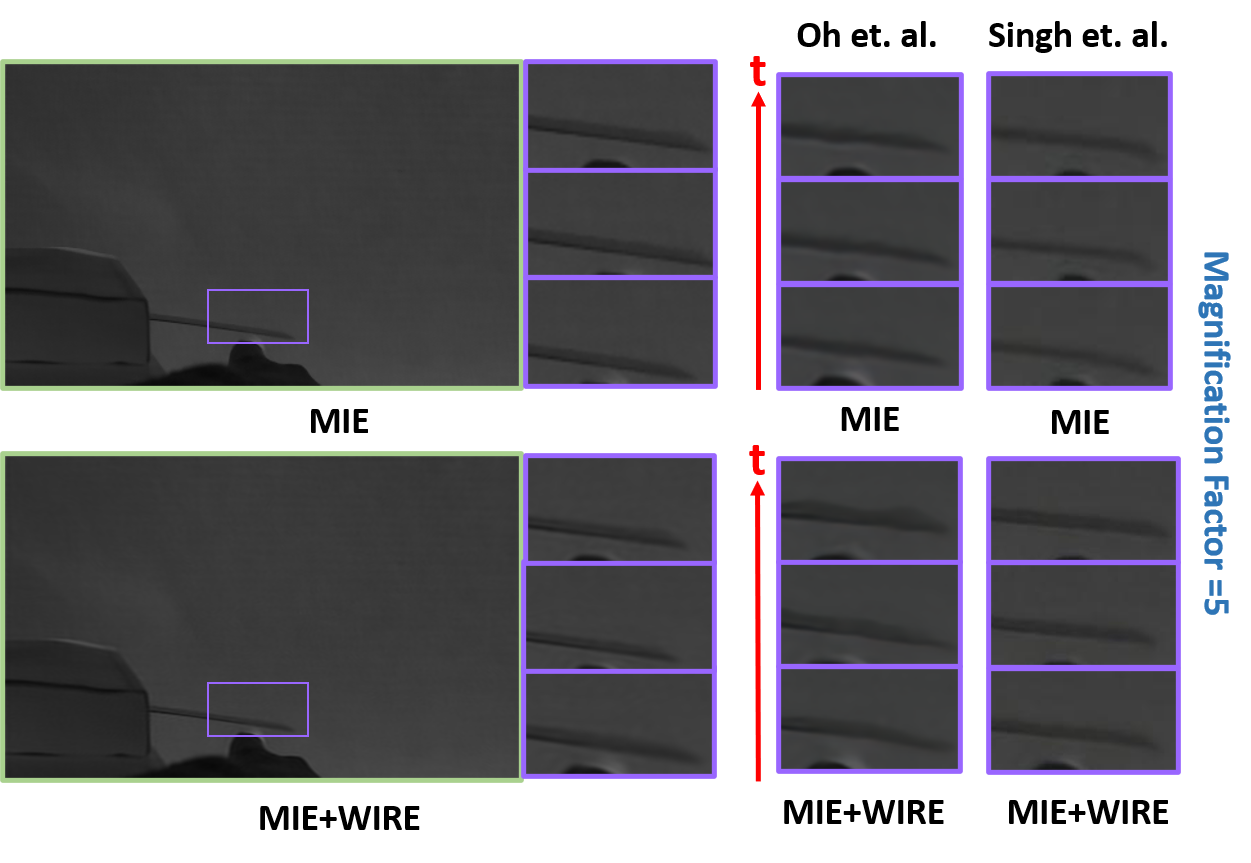}
    \caption{Temporal evolution of motion in the `Short Ruler' scene.}
    \label{fig:fig14}
\end{figure}
In Fig.~\ref{fig:fig14}, we show the temporal evolution of motion in the ‘Short Ruler’ scene using different methods. In our supplemental video of `Short Ruler', it can also be observed the effectiveness of SpikeMM in the magnification of high-frequency micro-movements.

\section{Conclusion}
The SpikeMM introduced in this paper shows unprecedented potential in the field of high-speed micro-motion amplification. SpikeMM, leveraging the unique ability of spike cameras to capture temporal frequency domains, overcomes the challenges faced by traditional algorithms in high-speed scenarios due to motion blur. By integrating multi-level information extraction, spatial upsampling, and motion magnification modules, this algorithm offers a self-supervised approach adaptable to a wide range of scenarios, seamlessly integrating with high-performance super-resolution and motion magnification algorithms. Rigorous validation using scenes captured by spike cameras has substantiated the capacity of SpikeMM to accurately magnify motions in real-world high-frequency settings.

In the future, SpikeMM is expected to play a bigger role in several fields. For example, in mechanical fault detection, by amplifying the subtle vibrations in equipment operation, potential problems can be detected early and downtime and losses caused by sudden equipment failure can be avoided. In fluid mechanics research, SpikeMM can be used to observe and analyze the subtle movements of high-speed fluids to help optimize designs, reduce drag, and increase efficiency. In the medical field, SpikeMM can be used for dynamic monitoring of organs such as the heart and blood vessels to help doctors see physiological activity more clearly and make more accurate diagnoses and treatment decisions. In addition, SpikeMM can also be used for security monitoring, by amplifying small movements in surveillance videos, early detection of potential security threats, improve the efficiency and accuracy of security monitoring.



\bibliographystyle{IEEEtran}


 




\bibliography{SpikeMM}
\end{document}